\documentclass[11pt,a4paper]{article}
\usepackage[hyperref]{acl2021}
\usepackage{times}
\usepackage{latexsym}

\usepackage{microtype}

\aclfinalcopy 
\def\aclpaperid{3673} 

\usepackage{booktabs}
\usepackage{graphicx}
\usepackage{amsmath}
\usepackage{amssymb}
\usepackage{algorithm}
\usepackage[ruled,algo2e]{algorithm2e}
\usepackage{multirow}
\usepackage{bm}
\usepackage{amsfonts}
\usepackage{subcaption}
\usepackage{todonotes}
\usepackage{balance}
\usepackage{enumitem}
\usepackage{xcolor}
\usepackage{balance}
\usepackage{wrapfig}
\usepackage{relsize}
\usepackage{subcaption}
\usepackage{wrapfig}
\usepackage{graphics}
\usepackage{xr}
\usepackage{cleveref}

\title{Generating SOAP Notes from Doctor-Patient Conversations\\
Using Modular Summarization Techniques}

\author{Kundan Krishna, Sopan Khosla, Jeffrey Bigham, Zachary C. Lipton\\
  Carnegie Mellon University \\
  5000 Forbes Avenue \\
  Pittsburgh, PA \\
  \texttt{\{kundank,sopank,jbigham,zlipton\}@andrew.cmu.edu} \\
  }

\date{}

\begin{document}

\maketitle

\begin{abstract}
Following each patient visit, 
physicians draft long semi-structured
clinical summaries called SOAP notes.
While invaluable to clinicians and researchers,
creating digital SOAP notes is burdensome,
contributing to physician burnout.
In this paper, we introduce 
the first complete pipelines
to leverage deep summarization models
to generate these notes 
based on transcripts of conversations 
between physicians and patients.
After exploring a spectrum of methods
across the extractive-abstractive spectrum,
we propose {\sc Cluster2Sent},
an algorithm that
(i) extracts important utterances
relevant to each summary section;
(ii) clusters together related utterances;
and then (iii) generates 
one summary sentence
per cluster. 
{\sc Cluster2Sent} outperforms
its purely abstractive counterpart
by 8 ROUGE-1 points,
and produces significantly more 
factual and coherent sentences 
as assessed by expert human evaluators.
For reproducibility, 
we demonstrate similar benefits
on the publicly available AMI dataset. 
Our results speak to the benefits
of structuring summaries into sections 
and annotating supporting evidence 
when constructing summarization corpora.

\end{abstract}

\maketitle

\section{Introduction}
\label{sec:intro}

\begin{figure*}[ht]
    \includegraphics[width=\textwidth]{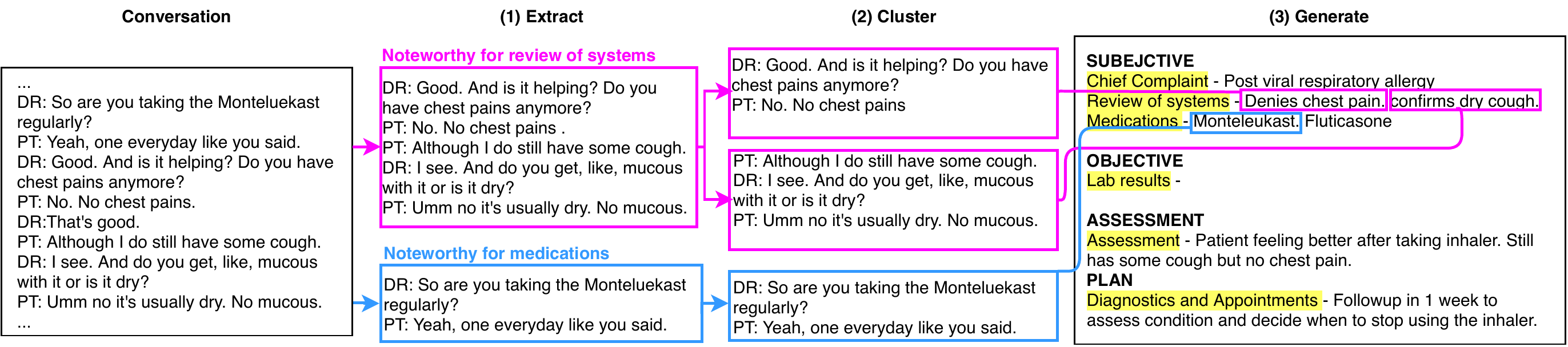}
    \caption{Workflow of our best performing approach 
    involving extraction and clustering 
    of noteworthy conversation utterances  
    followed by abstractive summarization 
    of each cluster (fictitious data)}
    \label{fig:intro_figure}
\end{figure*}

Electronic health records (EHR) 
play a crucial role in patient care.
However, populating them can
take as much time
as attending to patients 
\citep{sinsky2016allocation}
and constitutes a major cause 
of physician burnout 
\citep{kumar2020physician}.
In particular, doctors document 
patient encounters with SOAP notes, 
semi-structured written accounts 
containing four sections:
(S)ubjective information reported by the patient;
(O)bjective observations, e.g., lab results;
(A)ssessments made by the doctor 
(typically, the diagnosis);
and a (P)lan for future care, 
including diagnostic tests,
medications, and treatments.
Sections can be subdivided into 
$15$ subsections.

In a parallel development, 
patients increasingly record their doctor's visits,
either in lieu of taking notes or 
to share with a family member.
A budding line of research 
has sought to leverage transcripts 
of these clinical conversations 
both to provide insights to patients 
and to extract structured data 
to be entered into EHRs
\citep{liu2019fast, schloss2020towards, krishna2021extracting}.

In this paper, we introduce
the first end-to-end methods 
for generating whole SOAP notes 
based on clinical conversations.
Our work builds on a unique corpus,
developed in collaboration with 
\emph{Abridge AI, Inc.}\footnote{\url{http://abridge.com}}),
that consists of thousands of transcripts
of recorded clinical conversations 
together with associated SOAP notes 
drafted by a work force 
trained in the official style 
of SOAP note documentation.
On one hand, this task is much harder
than traditional summarization benchmarks,
in part, because SOAP notes are longer 
(320 words on average) 
than summaries in popular datasets like
CNN/Dailymail~\citep{nallapati2016abstractive}, 
Newsroom~\citep{grusky2018newsroom},
and SamSum~\citep{gliwa2019samsum} 
(55, 27, and 24 words on average). 
On the other hand, our dataset 
offers useful structure:
(i) segmentation of each SOAP note
into subsections;
and (ii) a set of \emph{supporting utterances} 
that provide evidence 
for each sentence in the SOAP note.
Exploiting this structure,
our methods outperform appropriate baselines.

Our first methodological contribution 
is to propose a spectrum of methods,
for decomposing summarizaton tasks
into \emph{extractive} 
and \emph{abstractive} subtasks.
Starting from a straightforward 
sequence-to-sequence model,
our methods shift progressively more work 
from the abstractive to the extractive component:
(i) {\sc Conv2Note:} 
the \emph{extractive} module does nothing,
placing the full burden of summarization 
on an end-to-end \emph{abstractive} module.
(ii) {\sc Ext2Note:} the extractive module selects 
all utterances that are \textit{noteworthy} 
(i.e., likely to be marked as supporting utterances 
for \emph{at least one} SOAP note sentence),
and the decoder is conditioned only on these utterances;
(iii) {\sc Ext2Sec:}  the extractive module 
extracts per-subsection \textit{noteworthy} utterances 
and the decoder generates each subsection,
conditioned only on the corresponding utterances;
(iv) {\sc Cluster2Sent:} the extractive module 
not only extracts per-subsection noteworthy utterances
but clusters together those likely to support 
the same SOAP sentence---here,
the decoder produces a single sentence at a time,
each conditioned upon a single cluster of utterances 
and a token indicating the SOAP subsection.
We see consistent benefits 
as we move from approach (i) through (iv).

Both to demonstrate the generality of our methods
and to provide a reproducible benchmark,
we conduct parallel experiments 
on the (publicly available) 
AMI corpus~\citep{carletta2007unleashing}\footnote{Our code
and trained models for the AMI dataset:
\url{https://github.com/acmi-lab/modular-summarization}}
Like our medical conversations dataset,
the AMI corpus exhibits 
section-structured summaries
and contains annotations 
that link summary sentences
to corresponding supporting utterances.
Our experiments with AMI data
show the same trends,
favoring pipelines that demand
more from the extractive component.
These results speak to the wider usefulness
of our proposed approaches,
{\sc Ext2Sec} and {\sc Cluster2Sent},
whenever section-structured summaries 
and annotated evidence utterances are available.

Our best performing model,
{\sc Cluster2Sent} (Figure~\ref{fig:intro_figure}),
demands the most of the extractive module,
requiring that it both select and group
each subsection's noteworthy utterances.
Interestingly, we observe that given
oracle (per-subsection) noteworthy utterances,
a simple proximity-based clustering heuristic 
leads to similar performance on SOAP note generation
as we obtain when using ground-truth clusters---even 
though the ground truth noteworthy utterances 
are not always localized.
Applied with predicted noteworthy utterances and clusters,
this approach achieves the highest ROUGE scores
and produces the most useful
(factual, coherent, and non-repetitive) sentences
as rated by human experts. 
As an additional benefit of this approach,
due to the smaller lengths 
of the input and output sequences involved, 
we can feasibly train large transformer-based 
abstractive summarization models (e.g., T5), 
whose memory requirements grow 
quadratically with sequence length.
Additionally, our approach localizes the precise utterances
upon which each SOAP note sentence depends,
enabling physicians to verify 
the correctness of each sentence
and potentially to improve the draft 
by highlighting the correct noteworthy utterances
(versus revising the text directly).

In summary, we contribute the following: 
\begin{itemize}
    \item The first pipeline for drafting entire SOAP notes 
    from doctor-patient conversations.
    \item A new collection of extractive-abstractive approaches 
    for generating long section-segmented summaries of conversations, 
    including new methods that leverage annotations attributing 
    summary sentences to conversation utterances.
    \item A rigorous quantitative evaluation 
    of our proposed models and appropriate baselines
    for both the extractive and abstractive components, 
    including sensitivity of the pipeline to simulated ASR errors.
    \item A detailed human study to evaluate 
    the factuality and quality of generated SOAP notes, 
    and qualitative error analysis.
\end{itemize}

\section{Related Work}

Summarization is a well-studied problem 
in NLP~\citep{nenkova2011automatic}. 
While early works focused on simply 
extracting important content from a document 
\citep{erkan2004lexrank, wong2008extractive},  
later approaches attempted to paraphrase the content
into new sentences (abstractive summarization)
\citep{filippova2010multi,berg2011jointly,wang2013domain}.
Following the development of neural sequence models 
\citep{sutskever2014sequence}, 
more research focuses on neural generation 
of abstractive summaries
\citep{nallapati2016abstractive,see2017get, celikyilmaz2018deep}. 
While many papers summarize news articles, 
others summarize conversations,
in business meetings \citep{wang2013domain,zhu2020end}, 
customer service \citep{liu2019automatic},
and tourist information center~\cite{yuan2019abstractive} contexts.

In the space of two-step extractive-abstractive summarization 
approaches, \citet{subramanian2019extractive} summarize 
scientific papers by first extracting sentences from
it and then abstractively summarizing them. 
\citet{chen2018fast} extract important sentences 
from the input and then paraphrase each of them
to generate the abstractive summary. 
While they assume that each summary sentence 
is supported by exactly one source sentence,
in our medical conversations, 
many summary sentences synthesize content
spread across multiple dialogue turns
(e.g., a series of questions and answers).

Past work on abstractive summarization of medical conversations 
has focused on summarizing patient-nurse conversations
with goals including capturing symptoms of interest \citep{liu2019topic}
and past medical history \citep{joshi2020dr}.
These tasks are respectively similar to generating  
the \textit{review of systems} 
and \textit{past medical history} 
subsections of a SOAP note. 
In contrast, we aim to generate a full-length SOAP note 
containing up to $15$ subsections,
and propose methods to 
address this challenge by extracting supporting context 
for smaller parts and generating them independently.

\section{Dataset}

We use two different datasets in this work. 
The primary medical dataset,
developed through a collaboration with Abridge AI,
consists of doctor-patient conversations 
with annotated SOAP notes. 
Additionally, we evaluate our summarization methods 
on the AMI dataset \citep{carletta2007unleashing},
comprised of business meeting transcripts and their summaries.

\subsection{Medical dataset}
Our work builds on a unique resource:
a corpus consisting of thousands of recorded 
English-language clinical conversations, 
with associated SOAP notes created by a work force 
trained in SOAP note documentation standards.
Our dataset consists of transcripts 
from real-life patient-physician visits 
from which sensitive information such 
as names have been de-identified.
The full medical dataset consists of $6862$ visits 
consisting of $2732$ cardiologist visits,
$2731$ visits for family medicine, 
$989$ interventional cardiologist visits,
and $410$ internist visits.
Owing to the sensitive nature of the data,
we cannot share it publicly 
(an occupational hazard of research 
on machine learning for healthcare).

For each visit, our dataset contains 
a human-generated transcript of the conversation.
The transcript is segmented
into utterances, each annotated
with a timestamp and speaker ID.
The average conversation lasts $9.43$ minutes 
and consists of around $1.5\text{k}$ words 
(Appendix Figure~\ref{fig:datasetstats}).
Associated with each conversation,
we have a human-drafted SOAP note 
created by trained, professional annotators. 
The annotators who created the SOAP notes
worked in either clinical transcription, 
billing, or related documentation-related departments, 
but were not necessarily professional medical scribes.
The dataset is divided into train,
validation and test splits of size 
$5770$, $500$ and $592$, respectively.

Our annotated SOAP notes contain (up to) $15$ subsections,
each of which may contain multiple sentences. 
The subsections vary in length.
The \textit{Allergies} subsections is most often empty,
while the  \textit{Assessment} subsection contains
$5.16$ sentences on average 
(Table~\ref{tab:subsection_lengths}).
The average SOAP note contains $27.47$ sentences.
The different subsections also 
differ in the style of writing.
The \textit{Medications} subsection 
usually consists of bulleted names 
of medicines and their dosages, 
while the \textit{Assessment} subsection
typically contains full sentences.
On average, the fraction of novel 
(i.e., not present in the conversation) 
unigrams, bigrams, and trigrams,
in each SOAP note are 
$24.09\%$, $67.79\%$ and $85.22\%$, respectively.

\begin{table}[t]
    \centering
\resizebox{\linewidth}{!}{
    \begin{tabular}{lc}
    \toprule
\textbf{Subsection} &	\textbf{Mean length} \\
\midrule
Family Medical History  & 0.23 \\	
Past Surgical History 	& 0.58  \\
Review of Systems 	& 3.65 \\
Chief Complaint	& 2.17 \\
Miscellaneous	& 2.81 \\
Allergies	& 0.06 \\
Past Medical History 	& 2.93 \\	
Social History	& 0.27 \\	
Medications	& 3.74 \\
\midrule
 Immunizations	& 0.11 \\
Laboratory and Imaging Results 	& 2.27 \\
\midrule
Assessment	& 5.16 \\
\midrule
 Diagnostics and Appointments 	& 1.65 \\
Prescriptions and Therapeutics 	& 1.75\\
\midrule
Healthcare Complaints	& 0.09\\
\bottomrule
    \end{tabular}
}
    \caption{Average number of sentences in different SOAP note subsections 
    grouped by parent sections (\textit{Subjective}, \textit{Objective}, \textit{Assessment}, \textit{Plan}, \textit{Others} resp.)}
    \label{tab:subsection_lengths}
    \vspace{-10px}
\end{table}

Each SOAP note sentence is also annotated with utterances 
from the conversation which provide evidence for that sentence. 
A SOAP note sentence can have 
one or more supporting utterances. 
On average, each SOAP sentence has $3.84$ supporting utterances,
but the mode is $1$ (Appendix Figure~\ref{fig:datasetstats}).
We refer to these utterances as \emph{noteworthy utterances} throughout this paper.
Throughout this work, we deal with 
the $15$ more granular subsections 
rather than the $4$ coarse sections of SOAP notes,
and thus for convenience, 
all further mentions of \emph{section}
technically denote a SOAP \emph{subsection}.

\subsection{AMI dataset}
The AMI dataset is a collection of 138 business meetings, 
each with 4 participants with various roles 
(e.g., marketing expert, product manager, etc.). 
Each meeting transcript comes 
with an associated abstractive summary 
that is divided into four sections---\textit{abstract}, 
\textit{decisions}, \textit{actions}, and \textit{problems}. 
Each conversation also has 
an associated extractive summary, 
and there are additional annotations 
linking the utterances in the extractive summary 
to sentences in the abstractive summary. 
For any given sentence in the abstractive summary,
we refer to the linked set of utterances 
in the extractive summary as its \emph{noteworthy utterances}.
We note that 7.9\% of the abstractive summary sentences 
have no annotated noteworthy utterances.
To simplify the analysis, 
we remove these sentences from 
summaries in the training,
validation, and test splits.

\section{Methods}
\label{sec:solution_approach}

We investigate the following four decompositions 
of the summarization problem 
into extractive and abstractive phases,
ordered from abstraction-heavy to extraction-heavy:
    {\sc Conv2Note} takes an end-to-end approach,
    generating the entire SOAP note 
    from the entire conversation in one shot. 
    {\sc Ext2Note} first predicts 
    all of the noteworthy utterances
    in the conversation 
    (without regard to the associated section)
    and then generates the entire SOAP note 
    in one shot from only those utterances. 
    {\sc Ext2Sec} extracts noteworthy utterances,
    while also predicting the section(s) 
    for which they are relevant, 
    and then generates each SOAP section separately 
    using only that section's predicted noteworthy utterances.    
    {\sc Cluster2Sent} attempts to group together
    the set of noteworthy utterances 
    associated with each summary sentence.
    Here, we cluster separately among each set 
    of section-specific noteworthy utterances
    and then generate each section one sentence at a time,
    conditioning each on the associated cluster of utterances.

Each of these pipelines leaves open many choices 
for specific models to employ for each subtask.
For the abstractive modules of {\sc Conv2Note} and {\sc Ext2Note},
we use a pointer-generator network. 
The abstractive modules of {\sc Ext2Sec} and {\sc Cluster2Sent},
which require conditioning on section 
are modeled using conditioned pointer-generator networks 
(described in Section~\ref{sec:modelarch}), 
and fine-tuned T5 models which condition 
on the section being generated 
by means of prepending it to the input.
T5 models could not be used in
the {\sc Conv2Note} and {\sc Ext2Note} settings 
because their high memory requirement 
for long inputs could not be accommodated 
even with 48GB of GPU memory.

For noteworthy utterance extraction, 
we primarily use a hierarchical LSTM model 
and a BERT-LSTM model as described in the next section. 
All models are configured to have a scalar output 
for binary classification in {\sc Ext2Note}, 
whereas for {\sc Ext2Sec} and {\sc Cluster2Sent}, 
they have multi-label output 
separately predicting noteworthiness for each section.
Note that the same utterance can be noteworthy 
with respect to multiple sections. 
We use the same trained utterance extraction models 
for both {\sc Ext2Sec} and {\sc Cluster2Sent}.

For the clustering module in {\sc Cluster2Sent}, 
we propose a heuristic 
that groups together 
any two supporting utterances 
that are \emph{close},
meaning they have less than or equal to 
$\tau$ utterances separating them, 
where $\tau$ is a hyperparameter.
This process is iterated, 
with the clusters growing in size 
by merging with other singletons or clusters,
until every pair of \emph{close} utterances 
have the same cluster membership.
The value of $\tau$
is tuned on the validation set. 
Since each cluster necessarily produces 
one sentence in the SOAP note, 
having too many or too few clusters 
can make the SOAP note 
too long or too short, respectively. 
Therefore, for any given value 
of the hyper-parameter $\tau$ and any given section, 
the prediction thresholds of the extractor 
are tuned on the validation set to produce 
approximately the same number of clusters
over the entire validation set 
as present in the ground truth for that section. 
Among ground truth clusters containing
multiple noteworthy utterances, 
$82\%$ are contiguous. 
In an experiment where the heuristic is used to cluster 
the \emph{oracle noteworthy utterances} for each section, 
and summaries are subsequently generated 
via the abstractive modules from {\sc Cluster2Sent}, 
ROUGE-1 and ROUGE-2 metrics deteriorate 
by less than $1$ point as compared 
to \emph{oracle} clusterings 
(Appendix Table~\ref{tab:rouge_scores_detailed}), 
demonstrating our heuristic's effectiveness.

\section{Model Architectures}
\label{sec:modelarch}

\vspace{5px}
\noindent \textbf{Pointer-Generator Network \quad}
We use the pointer-generator network 
introduced by \citet{see2017get} 
for {\sc Conv2Note} and {\sc Ext2Note}.
The model is a bidirectional LSTM-based 
encoder-decoder model with attention. 
It employs a pointer mechanism 
to copy tokens directly from the input
in addition to generating them 
by predicting generation probabilities
for the entire vocabulary. 
The model also computes the weights
that govern copying versus generating 
at each decoding timestep.

\vspace{5px}
\noindent \textbf{Section-conditioned Pointer-Generator Network \quad}
We modify the pointer-generator network 
for algorithms {\sc Ext2Sec} and {\sc Cluster2Sent},
to condition on the (sub)section 
of the summary to be generated.
The network uses a new lookup table 
to embed the section $z$ 
into an embedding $\bm{e}^z$. 
The section embedding is concatenated 
to each input word embedding fed into the encoder. 
The section embedding is also appended to the inputs 
of the decoder LSTM in the same fashion.

\vspace{5px}
\noindent \textbf{T5 \quad}
We use the recently released T5 model 
\citep{raffel2020exploring} as an abstractive module.
It is an encoder-decoder model,
where both encoder and decoder 
consist of a stack of transformer layers. 
The T5 model is pre-trained on 5 tasks, 
including summarization, translation etc. 
We use the pre-trained T5 model parameters 
and fine-tune it on our task dataset. 
For introducing the section-conditioning 
in {\sc Ext2Sec} and {\sc Cluster2Sent}, 
we simply add the name of the section 
being generated to the beginning of the input.

\vspace{5px}
\noindent \textbf{Hierarchical LSTM classifier(H-LSTM) \quad}
In this model, we first encode each utterance $u_i$ independently 
by passing its tokens through a bidirectional LSTM 
and mean-pooling their encoded representations 
to get the utterance representation $\bm{h}_i$.
We pass the sequence of utterance representations 
$\{\bm{h}_1,\bm{h}_2,...,\bm{h}_n\}$
through another bidirectional LSTM 
to get new utterance representations 
which incorporate neighboring contexts. 
These are then passed through a sigmoid activated linear layer 
to predict each utterance's probability of noteworthiness 
with respect to each section.

\vspace{5px}
\noindent \textbf{BERT-LSTM classifier(B-LSTM) \quad} 
In this model, tokens in the utterance $u_i$ 
are passed through a \texttt{BERT} encoder 
to obtain their contextualized representations,
which are mean-pooled to get the utterance representation $\bm{h}_i$.
The subsequent architecture exactly mirrors hierarchical LSTM,
and involves passing utterance representations 
through a bidirectional LSTM and linear layer to get output probabilities. 
BERT-LSTM is fine-tuned in an end-to-end manner.

\section{Experiments}
We first establish two baselines. 
{\sc RandomNote} randomly and uniformly 
samples a SOAP note from the training set 
and outputs it as the summary 
for any input conversation.
{\sc OracleExt} presents all the
ground truth noteworthy utterances (evidence)
from the conversation as the SOAP note
without any abstractive summarization. 
Thus, the {\sc OracleExt} baseline has the advantage 
of containing all the desired information
(e.g., names of medicines) from the conversation, 
but the disadvantage of not being expressed 
in the linguistic style of a SOAP note 
which leads to lower n-gram overlap. 
The opposite is true for the {\sc RandomNote} baseline.
Both baselines give similar performance 
and are outperformed by the simple {\sc Conv2Note} approach 
(Table~\ref{tab:methods_rouge}).

We train the abstractive modules for 
the 4 approaches described in Section~\ref{sec:solution_approach}
with the ground truth noteworthy utterances as inputs. 
To estimate an upper bound on the performance 
we can reasonably hope to achieve 
by improving our noteworthy utterance extractors, 
we test our models with oracle noteworthy utterances in the test set.
All algorithms relying on oracle noteworthy utterances
outperform {\sc Conv2Note}, and exhibit 
a monotonic and significant rise in ROUGE scores 
as we move towards the extraction-heavy end of the spectrum
(Table~\ref{tab:oracle_methods_rouge1})\footnote{ The character `-' represents GPU memory overflow}.

\begin{table*}[t!]
    \centering
    \begin{tabular}{l c c c c c c c}
    \toprule
    & \multicolumn{3}{c}{\textbf{Medical dataset}} & \hspace{10pt} & \multicolumn{3}{c}{\textbf{AMI corpus}} \\
    \midrule
    \textbf{Method} & \textbf{R-1} & \textbf{R-2} & \textbf{R-L} & & \textbf{R-1} & \textbf{R-2} & \textbf{R-L}  \\
    \midrule
        {\sc RandomNote} & 34.99 & 12.69 & 21.37  & & 42.47 & 11.55 & 21.47\\
        {\sc OracleExt} & 33.07 & 12.22 & 17.42  & & 39.97 & 11.17 & 20.91 \\   
        \midrule
        {\sc Conv2Note} (PG) & 49.56 & 25.68 & 32.87 & & 39.62 & 13.16 & 23.95\\
        \midrule
        {\sc Ext2Note} (PG + HLSTM) & 49.58	& 24.91 & 31.68 & & 21.28 &	7.06 &	15.96\\
        {\sc Ext2Note} (PG + BLSTM) & 50.50 & 25.4 & 31.93 & & 21.71 &	6.83 &	15.69\\
        {\sc Ext2Note} (T5-Small  + HLSTM) & - & - & - & & 40.48 &	13.82 &	24.64\\
        {\sc Ext2Note} (T5-Small  + BLSTM) & - & - & - & & 40.36 &	13.73 &	24.13\\        
        \midrule
        {\sc Ext2Sec} (PG + HLSTM) & 55.23 &	27.14 &	35.15 & & 43.75 &	15.25 &	23.46\\
        {\sc Ext2Sec} (PG + BLSTM) & 55.74 &	27.54 &	36.09 & & 40.48 &	15.61 &	23.31\\
        {\sc Ext2Sec} (T5-Small + HLSTM)& 55.77 &	28.64 &	37.50 & & 42.45 &	15.20 &	23.92\\
        {\sc Ext2Sec} (T5-Small + BLSTM)& 56.00 &	29.16 &	\textbf{38.38} & & 45.44 &	16.59 &	\textbf{26.14}\\
        \midrule
        {\sc Cluster2Sent} (PG + HLSTM) & 55.46	& 27.41 &	35.81 & & 46.19 &	16.64 &	24.29 \\
        {\sc Cluster2Sent} (PG + BLSTM) & 55.60 &	27.68 &	36.29 & & 42.31 &	15.92 &	23.51\\
        {\sc Cluster2Sent} (T5-Small + HLSTM)& 56.88 &	28.63 &	36.78& & 45.10 &	15.06 &	23.52\\
        {\sc Cluster2Sent} (T5-Small + BLSTM)& 57.14 &	29.11 &	37.43 & & 42.38	& 15.36 &	23.9\\
        {\sc Cluster2Sent} (T5-Base + HLSTM)& 57.27 &	29.10 &	37.38 & & \textbf{50.52} &	17.56 &	24.89\\        
        {\sc Cluster2Sent} (T5-Base + BLSTM)& \textbf{57.51}	& \textbf{29.56} &	38.06 & & 45.91	& \textbf{17.70} &	25.24\\     
    \bottomrule
    \end{tabular}
    \caption{ROUGE scores achieved by different methods on the two datasets}
    \label{tab:methods_rouge}
    \vspace{-0.3cm}
\end{table*}

\begin{table}[t!]
    \centering
\resizebox{\linewidth}{!}{
    \begin{tabular}{l c c c c c }
    \toprule
    & \multicolumn{2}{c}{\textbf{Medical dataset}} &  \multicolumn{2}{c}{\textbf{AMI corpus}} \\
    \midrule    
    Method & \small{PG} & \small{T5-Small} & \small{PG} & \small{T5-Small} \\    
    \midrule
    {\sc Ext2Note} & 52.95 & - & 20.44 & 41.10 \\
    {\sc Ext2Sec} & 61.00 & 62.37 & 43.32 & 46.85 \\
    {\sc Cluster2Sent} & 63.63 & 66.50 & 51.86 & 54.23 \\
    \bottomrule
    \end{tabular}
}
    \caption{ROUGE-1 achieved on test set when using 
    the abstractive models with oracle noteworthy utterances and clusters (more results with oracle in the Appendix)}
    \label{tab:oracle_methods_rouge1}
    \vspace{-0.3cm}
\end{table}

For predicting noteworthy utterances, 
we use two baselines:
(i) logistic regression on TF-IDF utterance representations;
and (ii) a model with a bidirectional LSTM 
to compute token-averaged utterance representations, 
followed by a linear classification layer.
These two models make the predictions
for each utterance independent of others. 
In contrast, we also use models 
which incorporate context 
from neighboring utterances:
(a) a hierarchical LSTM;
and (b) a BERT-LSTM model 
as described in Section~\ref{sec:modelarch}. 
The latter two methods perform much better
(Table~\ref{tab:classification_results}), 
demonstrating the benefit 
of incorporating neighboring context, 
with BERT-LSTM performing the best 
(see Appendix Table~\ref{tab:extractor_performance_sectionwise}
for section-wise performance).

Using predicted noteworthy utterances and clusters
instead of oracle ones leads to a drop in ROUGE scores, 
but the performance of {\sc Ext2Sec} and {\sc Cluster2Sent} 
is still better than {\sc Conv2Note} (Table~\ref{tab:methods_rouge}).
For the medical dataset, using a BERT-LSTM extractor 
leads to the best performance, 
with {\sc Cluster2Sent} outperforming {\sc Conv2Note} 
by about $8$ points in ROUGE-1 
(see Appendix Table~\ref{tab:sectionwise_results} 
for section-wise performance).
Interestingly, the T5-Small variant 
achieves similar performance to T5-Base, 
despite being only about a quarter of the latter's size.

\vspace{5px}
\noindent \textbf{Performance on AMI dataset \quad}
We see a similar trend in the ROUGE scores 
when applying these methods on the AMI dataset. 
One exception is the poor performance 
of pointer-generator based {\sc Ext2Note},
which excessively repeated sentences 
despite using a high coverage loss coefficient. 
There is a larger gap between the performance 
of the T5-Small and T5-Base abstractive models on this dataset. 
As an extractor, the performance of BERT-LSTM 
is again better than HLSTM 
(Table~\ref{tab:classification_results}), 
but when used in tandem with the abstractive module, 
ROUGE scores achieved by the overall pipeline 
do not always follow the same order.
We also observe that the clustering heuristic
does not work as well on this dataset.
Specifically, tuning the thresholds of the extractive model,
while fixing the clustering threshold $\tau$ 
gave worse results on this dataset. 
Tuning the thresholds independent of the clusters 
performed  better.
However, the best method still outperforms 
{\sc Conv2Note} by about 11 ROUGE-1 points 
(Table~\ref{tab:methods_rouge}).

\begin{table}[t]
    \centering
\resizebox{\linewidth}{!}{    
    \begin{tabular}{l c c c}
    \toprule
    \textbf{Method} & \textbf{R-1} & \textbf{R-2} & \textbf{R-L} \\
    \midrule
    \multicolumn{4}{c}{\textbf{Train on clean data + Test on data with 10\% error rate}}\\
    \midrule
        {\sc Conv2Note}(PG) & 46.52	& 22.60 & 30.45\\
        {\sc Cluster2Sent}(PG + BLS) & 51.84 &	23.74 &	32.94\\
        {\sc Cluster2Sent}(T5-Base+ BLS) & 54.88 & 26.65 & 35.88\\
        \midrule
        \multicolumn{4}{c}{\textbf{Train and test on data with 10\% error rate}}\\
        \midrule  
        {\sc Conv2Note}(PG) & 48.85 &	24.85 &	31.27\\
        {\sc Cluster2Sent}(PG + BLS) &  54.68	& 26.59 &	35.70\\
        {\sc Cluster2Sent}(T5-Base+ BLS) &  56.35 &	28.50 &	37.04\\
        \midrule        \multicolumn{4}{c}{\textbf{Train and test on data with 30\% error rate}}\\
        \midrule
        {\sc Conv2Note}(PG) & 45.16 &	22.26 &	30.14\\
        {\sc Cluster2Sent}(PG + BLS) &  53.69	& 25.88 & 35.12\\
        {\sc Cluster2Sent}(T5-Base+ BLS) &  55.90 &	27.73 &	36.06\\
    \bottomrule
    \end{tabular}
}
    \caption{Performance of models trained and tested on data with different simulated ASR error rates. BLS: BERT-LSTM}
    \label{tab:asr_errors}
\end{table}

\vspace{5px}
\noindent \textbf{Performance with ASR errors \quad}
In the absence of human-generated transcripts of conversations, 
Automatic Speech Recognition (ASR) techniques can be used 
to transcribe the conversations for use by our models. 
To account for ASR errors, we artificially added errors 
in transcripts of the medical dataset
by randomly selecting some percentage of the words 
and replacing them with phonetically similar words 
using RefinedSoundEx~\citep{refined_soundex} (details in the Appendix).
Models trained on clean dataset perform worse 
on a $10\%$ corrupted test dataset 
(Table~\ref{tab:asr_errors}). 
Since ASR errors lead to replacement of a correct word 
by only a small set of phonetically similar words,
there is still some information 
indicating the original word 
that can be used by the models.
When we train our models on data
corrupted at the 10\% ASR error rate,
our models recover much of the performance drop
(Table~\ref{tab:asr_errors}). 
Notably when simulated ASR errors 
are dialed up to a 30\% error rate,
(both at train and test time)
we see a smaller performance drop 
for {\sc Cluster2Sent} 
as compared to {\sc Conv2Note}.

\begin{table}[t]
    \centering
\resizebox{\linewidth}{!}{
\begin{tabular}{l c c c c c c c}
\toprule
& \multicolumn{4}{c}{\textbf{Medical conversations}} & \hspace{1pt} & \multicolumn{2}{c}{\textbf{AMI corpus}}\\
\midrule
\textbf{Metric} &  \textbf{LR} &  \textbf{LS} & \textbf{HLS} & \textbf{BLS} &  & \textbf{HLS} & \textbf{BLS} \\
\midrule
Accuracy        & 96.0  &  96.1 & 96.5 & 96.5 & & 93.77 & 94.16 \\
Ma-AUC       & 78.1  &  79.3 & 90.0 & 90.5 & & 83.81 & 90.76 \\
Ma-F1        & 29.5  &  31.0 & 38.6 & 40.9 & & 19.95 & 33.08 \\
Mi-AUC       & 87.3  &  87.6 & 92.7 & 93.3 & & 93.21 & 94.90 \\
Mi-F1        & 31.2  &  32.9 & 39.6 & 41.1 & & 43.76 & 49.93 \\
\bottomrule
\end{tabular}
}
    \caption{Performance on multilabel classification of noteworthy utterances 
    with logistic regression(LR), LSTM(LS), 
    Hierarchical-LSTM(HLS) and BERT-LSTM(BLS). 
    Ma:macro-averaged. Mi:micro-averaged }
    \label{tab:classification_results}
    \vspace{-0.3cm}
\end{table}

\section{Qualitative Analysis}
\label{sec:qual_analysis}

\begin{figure*}[t!]
    \centering
    \includegraphics[width=\textwidth]{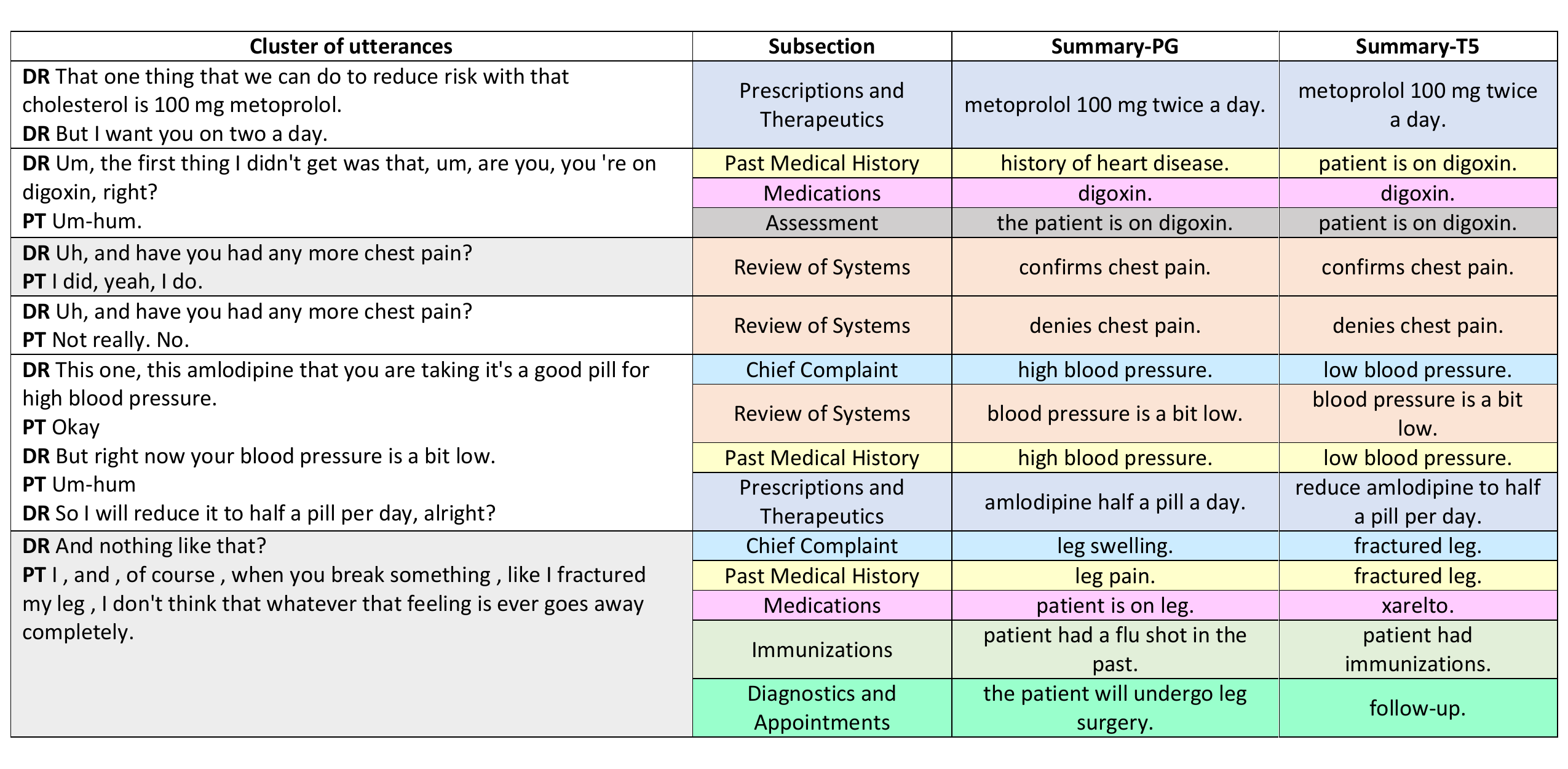}
    \vspace{-0.95cm}
    \caption{Noteworthy utterance clusters summarized in different ways 
    for different sections by the abstractive summarization modules of {\sc Cluster2Sent} 
    (utterances were slightly obfuscated for privacy reasons)}
    \label{fig:sample_entrywise_summaries}
    \vspace{-0.4cm}
\end{figure*}

The conditioned pointer-generator 
and T5 models used in {\sc Cluster2Sent} 
learn to place information 
regarding different topics 
in appropriate sections. 
Hence, given a cluster of supporting utterances, 
the models can generate different summaries 
for multiple sections 
(Figure~\ref{fig:sample_entrywise_summaries}). 
For example, given the same supporting utterances 
discussing the patient's usage 
of \textit{lisinopril} for low blood pressure, 
a model generates ``low blood pressure'' 
in the \textit{review of systems} section,
and ``lisinopril'' in \textit{medications} section.
We direct the reader 
to the appendix for examples of full-length 
generated SOAP notes.

Interestingly, when the abstractive model
is given a cluster of utterances that are 
not relevant to the section being generated,
the model sometimes outputs fabricated information 
relevant to that section
such as saying the patient is a non-smoker 
in \textit{social history},
or that the patient has taken a flu shot 
in \textit{immunizations} .
Hence, the quality of produced summaries 
heavily depends on the ability 
of the extractive step  
to classify the extracted utterances to the correct section. 
Another cause of false information
is the usage of pronouns in clusters 
without a mention of the referred entity. 
In such situations, T5 models frequently 
replace the pronoun with some arbitrary entity
(e.g. ``she'' with ``daughter'', compounds with ``haemoglobin'',
and medicines with ``lisinopril'').

Occasionally, the abstractive module produces 
new \emph{inferred} information 
that is not mentioned explicitly in the conversation.
In one instance, the model generated 
that the patient has a history of heart disease
conditioned on a cluster that mentioned 
he/she takes \emph{digoxin}, 
a popular medicine for heart disease. 
Similarly, the model can infer
past medical history of ``high cholesterol'' 
upon seeing  \emph{pravastatin} usage.
Such inferences can also lead to incorrect summaries,
e.g., when a doctor explained
that a patient has leaky heart valves, a model added a sentence 
to the \emph{diagnostics and appointments} section saying ``check valves''.

{\sc Cluster2Sent} summarizes localized regions
of the conversation \emph{independently}, 
which may lead to contradictions
in the SOAP note.
In one visit, the patient was asked about chest pain twice---once 
in the beginning to get to know his/her current state, 
and once as a question about how he/she felt 
just before experiencing a fall in the past.
This led to the model generating both 
that the patient denied chest pain 
as well as confirmed chest pain, without
clarifying that one statement was for the present
and another for the past.

\section{Human evaluation}
We asked trained human annotators to evaluate
generated SOAP notes for $45$ conversations. 
Every sentence in each SOAP note was labeled 
according to various quality dimensions such whether it was factually correct, 
incoherent, irrelevant, redundant, 
or placed under an inappropriate section.
The 
detailed statistics of annotations received for each quality dimension are provided in the Appendix. 
We also collected aggregate annotations 
for the comprehensiveness of each SOAP note 
and the extent to which it verbatim copied the transcript
on a 5-point Likert scale.

Human raters were presented 
with a web interface
showing the conversation, along with
a search feature to help them 
in looking up desired information.
The summaries generated by three methods ({\sc Conv2Note}(pointer-generator),  {\sc Cluster2Sent}(pointer-generator) and {\sc Cluster2Sent}(T5-base)) were presented in random order to hide their identities. 
For each sentence, we asked for
(i) Factual correctness of the sentence;
(ii) If the statement is simply repeating what has already been mentioned before;
(iii) If the statement is clinically irrelevant;
(iv) If the statement is incoherent (not understandable due to grammatical or semantic errors);
and (v) If the statement's topic does not match the section in which it is placed.
In addition, we asked two separate questions for rating the overall summary on a scale of 1-5 for its (i) comprehensiveness and (ii) extent of verbatim copying from conversation.
The human evaluation of the SOAP notes was done by workers who had also participated in the creation of the dataset of SOAP notes. Hence, they had already been extensively trained in the task of SOAP note creation, which gave them appropriate knowledge to judge the SOAP notes.

To quantify the performance among different methods, 
we consider a scenario where each generated SOAP note 
has to be post-edited by discarding undesirable sentences.
For a generated SOAP note, we define its \textit{yield} 
as the fraction of its total sentences that are not discarded. 
The sentences that are retained are those 
that are both factually correct
and were not labeled as either repetitive or incoherent. 
The human annotations show that both {\sc Cluster2Sent}-based methods tested 
produced a higher yield than the {\sc Conv2Note} baseline (p$<0.02$). 
T5-base performs better than conditioned pointer-generator 
as the abstractive module in {\sc Cluster2Sent} setting, 
producing significantly more yield 
(Table~\ref{tab:humaneval_results}). 
T5 also produces fewer incoherent sentences 
(Appendix Table~\ref{tab:humaneval_detailed_results}) 
likely due to its exposure to a large number 
of well-formed coherent sentences during pretraining.

We conducted an analogous human evaluation of summaries generated 
for all $20$ conversations in the test set of the AMI corpus, 
and saw a similar trend in the expected yield for different methods. 
Notably, for the AMI corpus, {\sc Conv2Note} produced 
a very high proportion of redundant sentences ($>0.5$) 
despite using the coverage loss, 
while the pointer-generator based {\sc Cluster2Sent} 
produced a high proportion of incoherent sentences 
(Appendix Table~\ref{tab:humaneval_detailed_results}).

\begin{table}[t!]
    \centering
\resizebox{\linewidth}{!}{
\begin{tabular}{l c c c c c c}
\toprule
& \multicolumn{3}{c}{\textbf{Medical conversations}} &  \multicolumn{3}{c}{\textbf{AMI corpus}}\\
\midrule
\textbf{Metric} &  \textbf{C2N} &  \textbf{C2S-P} & \textbf{C2S-T} &    \textbf{C2N} & \textbf{C2S-P} & \textbf{C2S-T}\\
\midrule
Length    & 21.2  &  28.2 & 28.4    & 20.7 &  17.9 & 19.05 \\
\%Yield     & 62.0  &  69.0 & 74.7    & 27.22 &  30.22 & 59.45 \\
Comp       & 2.44  &  2.42 & 2.76    & 2.30 &  2.55 & 3.75 \\
Copy       & 2.18  &  2.64 & 2.76    & 1.80 &  1.80 & 1.90 \\
\bottomrule
\end{tabular}
}
    \caption{Averages of different metrics for {\sc Conv2Note}(C2N), {\sc Cluster2Sent}  with pointer-generator (C2S-P) and T5-base (C2S-T). Comp:comprehensiveness, Copy:amount of copying. Length: number of sentences generated.}
    \label{tab:humaneval_results}
    \vspace{-0.3cm}
\end{table}

\section{Conclusion}

This paper represents the first attempt 
at generating full-length SOAP notes 
by summarizing transcripts 
of doctor-patient conversations.
We proposed a spectrum of extractive-abstractive 
summarization methods that leverage: 
(i) section-structured form of the SOAP notes
and (ii) linked conversation utterances associated with every SOAP note sentence. 
The proposed methods perform better than a fully abstractive approach and standard extractive-abstractive approaches that do not take advantage of these annotations.
We demonstrate the wider applicability of proposed approaches by showing similar results on the public AMI corpus which has similar annotations and structure.
Our work demonstrates the benefits of creating section-structured 
summaries (when feasible) and collecting evidence for each summary sentence
when creating any new summarization dataset.

\section*{Ethics Statement}

The methods proposed in this work to 
generate SOAP notes involve neural 
models that sometimes generate 
factually incorrect text~\cite{maynez2020faithfulness}.
The detection and correction of such factual errors in automatically generated summaries is an active area of research~\cite{cao2018faithful,zhang-etal-2020,dong2020multi}.
We emphasize that the methods are intended to be 
used with supervision from a
medical practitioner who 
can check for factual errors
and edit the the generated SOAP note if needed.
We have estimated the frequency of such factual errors (Appendix Table~\ref{tab:humaneval_detailed_results}) and characterized multiple types of errors seen in generated 
SOAP notes in Section~\ref{sec:qual_analysis}, 
for which the medical practitioners should remain vigilant.
For example, there is a bias to incorrectly generate information that occur frequently in specific sections (e.g. ``patient took flu shot''),
and to replace pronouns with frequently seen entities (such as ``lisinopril'' for references to medicine).
All data used in this study was manually  de-identified before we accessed it.
Deploying the proposed methods does not require long-term storage of conversations.
After the corresponding SOAP notes are generated,
conversations can be discarded. 
Hence, we do not anticipate any additional privacy risks 
from using the proposed methods.

\section*{Acknowledgements}
This work was funded by the Center for Machine Learning and Health in a joint venture between UPMC and Carnegie Mellon University.
We gratefully acknowledge support from Abridge AI, Inc.
for creating the dataset of SOAP notes and providing 
human resources for evaluation.

\bibliographystyle{acl_natbib}
\typeout{}
\bibliography{main.bib}

\begin{thebibliography}{32}
\expandafter\ifx\csname natexlab\endcsname\relax\def\natexlab#1{#1}\fi

\bibitem[{Berg-Kirkpatrick et~al.(2011)Berg-Kirkpatrick, Gillick, and
  Klein}]{berg2011jointly}
Taylor Berg-Kirkpatrick, Dan Gillick, and Dan Klein. 2011.
\newblock Jointly learning to extract and compress.
\newblock In \emph{Proceedings of the 49th Annual Meeting of the Association
  for Computational Linguistics: Human Language Technologies-Volume 1}, pages
  481--490. Association for Computational Linguistics.

\bibitem[{Cao et~al.(2018)Cao, Wei, Li, and Li}]{cao2018faithful}
Ziqiang Cao, Furu Wei, Wenjie Li, and Sujian Li. 2018.
\newblock Faithful to the original: Fact aware neural abstractive
  summarization.
\newblock In \emph{Proceedings of the AAAI Conference on Artificial
  Intelligence}, volume~32.

\bibitem[{Carletta(2007)}]{carletta2007unleashing}
Jean Carletta. 2007.
\newblock Unleashing the killer corpus: experiences in creating the
  multi-everything ami meeting corpus.
\newblock \emph{Language Resources and Evaluation}, 41(2):181--190.

\bibitem[{Celikyilmaz et~al.(2018)Celikyilmaz, Bosselut, He, and
  Choi}]{celikyilmaz2018deep}
Asli Celikyilmaz, Antoine Bosselut, Xiaodong He, and Yejin Choi. 2018.
\newblock Deep communicating agents for abstractive summarization.
\newblock In \emph{Proceedings of the 2018 Conference of the North American
  Chapter of the Association for Computational Linguistics: Human Language
  Technologies, Volume 1 (Long Papers)}, pages 1662--1675.

\bibitem[{Chen and Bansal(2018)}]{chen2018fast}
Yen-Chun Chen and Mohit Bansal. 2018.
\newblock Fast abstractive summarization with reinforce-selected sentence
  rewriting.
\newblock In \emph{Proceedings of the 56th Annual Meeting of the Association
  for Computational Linguistics (Volume 1: Long Papers)}, pages 675--686.

\bibitem[{Commons()}]{refined_soundex}
Apache Commons.
\newblock \href
  {https://commons.apache.org/proper/commons-codec/apidocs/org/apache/commons/codec/language/RefinedSoundex.html}
  {Refinedsoundex}.

\bibitem[{Dong et~al.(2020)Dong, Wang, Gan, Cheng, Cheung, and
  Liu}]{dong2020multi}
Yue Dong, Shuohang Wang, Zhe Gan, Yu~Cheng, Jackie Chi~Kit Cheung, and Jingjing
  Liu. 2020.
\newblock Multi-fact correction in abstractive text summarization.
\newblock In \emph{Proceedings of the 2020 Conference on Empirical Methods in
  Natural Language Processing (EMNLP)}, pages 9320--9331.

\bibitem[{Erkan and Radev(2004)}]{erkan2004lexrank}
G{\"u}nes Erkan and Dragomir~R Radev. 2004.
\newblock Lexrank: Graph-based lexical centrality as salience in text
  summarization.
\newblock \emph{Journal of artificial intelligence research}, 22:457--479.

\bibitem[{Filippova(2010)}]{filippova2010multi}
Katja Filippova. 2010.
\newblock Multi-sentence compression: Finding shortest paths in word graphs.
\newblock In \emph{Proceedings of the 23rd international conference on
  computational linguistics}, pages 322--330. Association for Computational
  Linguistics.

\bibitem[{Gliwa et~al.(2019)Gliwa, Mochol, Biesek, and Wawer}]{gliwa2019samsum}
Bogdan Gliwa, Iwona Mochol, Maciej Biesek, and Aleksander Wawer. 2019.
\newblock Samsum corpus: A human-annotated dialogue dataset for abstractive
  summarization.
\newblock In \emph{Proceedings of the 2nd Workshop on New Frontiers in
  Summarization}, pages 70--79.

\bibitem[{Grusky et~al.(2018)Grusky, Naaman, and Artzi}]{grusky2018newsroom}
Max Grusky, Mor Naaman, and Yoav Artzi. 2018.
\newblock Newsroom: A dataset of 1.3 million summaries with diverse extractive
  strategies.
\newblock \emph{arXiv preprint arXiv:1804.11283}.

\bibitem[{Joshi et~al.(2020)Joshi, Katariya, Amatriain, and
  Kannan}]{joshi2020dr}
Anirudh Joshi, Namit Katariya, Xavier Amatriain, and Anitha Kannan. 2020.
\newblock Dr. summarize: Global summarization of medical dialogue by exploiting
  local structures.
\newblock \emph{arXiv preprint arXiv:2009.08666}.

\bibitem[{Krishna et~al.(2021)Krishna, Pavel, Schloss, Bigham, and
  Lipton}]{krishna2021extracting}
Kundan Krishna, Amy Pavel, Benjamin Schloss, Jeffrey~P Bigham, and Zachary~C
  Lipton. 2021.
\newblock Extracting structured data from physician-patient conversations by
  predicting noteworthy utterances.
\newblock In \emph{Explainable AI in Healthcare and Medicine}, pages 155--169.
  Springer.

\bibitem[{Kumar and Mezoff(2020)}]{kumar2020physician}
Gogi Kumar and Adam Mezoff. 2020.
\newblock Physician burnout at a children’s hospital: Incidence,
  interventions, and impact.
\newblock \emph{Pediatric Quality \& Safety}, 5(5).

\bibitem[{Liu et~al.(2019{\natexlab{a}})Liu, Wang, Xu, Li, and
  Ye}]{liu2019automatic}
Chunyi Liu, Peng Wang, Jiang Xu, Zang Li, and Jieping Ye. 2019{\natexlab{a}}.
\newblock Automatic dialogue summary generation for customer service.
\newblock In \emph{Proceedings of the 25th ACM SIGKDD International Conference
  on Knowledge Discovery \& Data Mining}, pages 1957--1965.

\bibitem[{Liu et~al.(2019{\natexlab{b}})Liu, Lim, Suhaimi, Tong, Ong, Ng,
  Guang, Macdonald, Ramasamy, Krishnaswamy et~al.}]{liu2019fast}
Zhengyuan Liu, Jia Hui~Hazel Lim, Nur Farah~Ain Suhaimi, Shao~Chuen Tong,
  Sharon Ong, Angela Ng, Sheldon Lee~Shao Guang, Michael~Ross Macdonald,
  Savitha Ramasamy, Pavitra Krishnaswamy, et~al. 2019{\natexlab{b}}.
\newblock Fast prototyping a dialogue comprehension system for nurse-patient
  conversations on symptom monitoring.
\newblock In \emph{NAACL-HLT (2)}.

\bibitem[{Liu et~al.(2019{\natexlab{c}})Liu, Ng, Lee, Aw, and
  Chen}]{liu2019topic}
Zhengyuan Liu, Angela Ng, Sheldon Lee, Ai~Ti Aw, and Nancy~F Chen.
  2019{\natexlab{c}}.
\newblock Topic-aware pointer-generator networks for summarizing spoken
  conversations.
\newblock \emph{arXiv preprint arXiv:1910.01335}.

\bibitem[{Maynez et~al.(2020)Maynez, Narayan, Bohnet, and
  McDonald}]{maynez2020faithfulness}
Joshua Maynez, Shashi Narayan, Bernd Bohnet, and Ryan McDonald. 2020.
\newblock On faithfulness and factuality in abstractive summarization.
\newblock In \emph{Proceedings of the 58th Annual Meeting of the Association
  for Computational Linguistics}, pages 1906--1919.

\bibitem[{Nallapati et~al.(2016)Nallapati, Zhou, dos Santos, Gul{\c{c}}ehre,
  and Xiang}]{nallapati2016abstractive}
Ramesh Nallapati, Bowen Zhou, Cicero dos Santos, {\c{C}}a{\u{g}}lar
  Gul{\c{c}}ehre, and Bing Xiang. 2016.
\newblock Abstractive text summarization using sequence-to-sequence rnns and
  beyond.
\newblock In \emph{Proceedings of The 20th SIGNLL Conference on Computational
  Natural Language Learning}, pages 280--290.

\bibitem[{Nenkova et~al.(2011)Nenkova, McKeown et~al.}]{nenkova2011automatic}
Ani Nenkova, Kathleen McKeown, et~al. 2011.
\newblock Automatic summarization.
\newblock \emph{Foundations and Trends{\textregistered} in Information
  Retrieval}, 5(2--3):103--233.

\bibitem[{Raffel et~al.(2020)Raffel, Shazeer, Roberts, Lee, Narang, Matena,
  Zhou, Li, and Liu}]{raffel2020exploring}
Colin Raffel, Noam Shazeer, Adam Roberts, Katherine Lee, Sharan Narang, Michael
  Matena, Yanqi Zhou, Wei Li, and Peter~J Liu. 2020.
\newblock Exploring the limits of transfer learning with a unified text-to-text
  transformer.
\newblock \emph{Journal of Machine Learning Research}, 21(140):1--67.

\bibitem[{Schloss and Konam(2020)}]{schloss2020towards}
Benjamin Schloss and Sandeep Konam. 2020.
\newblock Towards an automated soap note: Classifying utterances from medical
  conversations.
\newblock In \emph{Machine Learning for Healthcare Conference}, pages 610--631.
  PMLR.

\bibitem[{See et~al.(2017)See, Liu, and Manning}]{see2017get}
Abigail See, Peter~J Liu, and Christopher~D Manning. 2017.
\newblock Get to the point: Summarization with pointer-generator networks.
\newblock In \emph{Proceedings of the 55th Annual Meeting of the Association
  for Computational Linguistics (Volume 1: Long Papers)}, pages 1073--1083.

\bibitem[{Sinsky et~al.(2016)Sinsky, Colligan, Li, Prgomet, Reynolds, Goeders,
  Westbrook, Tutty, and Blike}]{sinsky2016allocation}
Christine Sinsky, Lacey Colligan, Ling Li, Mirela Prgomet, Sam Reynolds,
  Lindsey Goeders, Johanna Westbrook, Michael Tutty, and George Blike. 2016.
\newblock Allocation of physician time in ambulatory practice: a time and
  motion study in 4 specialties.
\newblock \emph{Annals of internal medicine}.

\bibitem[{Subramanian et~al.(2019)Subramanian, Li, Pilault, and
  Pal}]{subramanian2019extractive}
Sandeep Subramanian, Raymond Li, Jonathan Pilault, and Christopher Pal. 2019.
\newblock On extractive and abstractive neural document summarization with
  transformer language models.
\newblock \emph{arXiv preprint arXiv:1909.03186}.

\bibitem[{Sutskever et~al.(2014)Sutskever, Vinyals, and
  Le}]{sutskever2014sequence}
Ilya Sutskever, Oriol Vinyals, and Quoc~V Le. 2014.
\newblock Sequence to sequence learning with neural networks.
\newblock In \emph{Advances in neural information processing systems}, pages
  3104--3112.

\bibitem[{Tu et~al.(2016)Tu, Lu, Liu, Liu, and Li}]{tu2016modeling}
Zhaopeng Tu, Zhengdong Lu, Yang Liu, Xiaohua Liu, and Hang Li. 2016.
\newblock Modeling coverage for neural machine translation.
\newblock In \emph{ACL}, pages 76--85.

\bibitem[{Wang and Cardie(2013)}]{wang2013domain}
Lu~Wang and Claire Cardie. 2013.
\newblock Domain-independent abstract generation for focused meeting
  summarization.
\newblock In \emph{Proceedings of the 51st Annual Meeting of the Association
  for Computational Linguistics (Volume 1: Long Papers)}, pages 1395--1405.

\bibitem[{Wong et~al.(2008)Wong, Wu, and Li}]{wong2008extractive}
Kam-Fai Wong, Mingli Wu, and Wenjie Li. 2008.
\newblock Extractive summarization using supervised and semi-supervised
  learning.
\newblock In \emph{Proceedings of the 22nd international conference on
  computational linguistics (Coling 2008)}, pages 985--992.

\bibitem[{Yuan and Yu(2019)}]{yuan2019abstractive}
Lin Yuan and Zhou Yu. 2019.
\newblock Abstractive dialog summarization with semantic scaffolds.
\newblock \emph{arXiv preprint arXiv:1910.00825}.

\bibitem[{Zhang et~al.(2020)Zhang, Merck, Tsai, Manning, and
  Langlotz}]{zhang-etal-2020}
Yuhao Zhang, Derek Merck, Emily Tsai, Christopher~D. Manning, and Curtis
  Langlotz. 2020.
\newblock Optimizing the factual correctness of a summary: A study of
  summarizing radiology reports.
\newblock In \emph{Proceedings of the 58th Annual Meeting of the Association
  for Computational Linguistics}, pages 5108--5120.

\bibitem[{Zhu et~al.(2020)Zhu, Xu, Zeng, and Huang}]{zhu2020end}
Chenguang Zhu, Ruochen Xu, Michael Zeng, and Xuedong Huang. 2020.
\newblock End-to-end abstractive summarization for meetings.
\newblock \emph{arXiv preprint arXiv:2004.02016}.

\end{thebibliography}

\setcounter{figure}{0}
\renewcommand{\thefigure}{A\arabic{figure}}

\setcounter{table}{0}
\renewcommand{\thetable}{A\arabic{table}}
\newpage

\newpage
\vspace{10pt}
\includegraphics[]{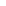}
\newpage

\appendix

\section*{Appendix}

\begin{table}[b!]
    \centering
    \begin{tabular}{l c c c}
    \toprule
    \textbf{Copy mechanism} & \textbf{R-1} & \textbf{R-2} & \textbf{R-L}  \\
    \midrule
    Present  & 63.63 & 35.62 & 48.85\\
    Absent  & 61.92 & 34.37 & 47.86\\
    \bottomrule
    \end{tabular}
    \caption{Impact of copy mechanism in peformance of a pointer-generator model on medical dataset in {\sc Cluster2Sent} using oracle noteworthy utterance clusters}
    \label{tab:mustusecopy}
\end{table}

\begin{table}[b!]
    \centering
    \begin{tabular}{l c c c}
    \toprule
    \textbf{Model} & \textbf{R-1} & \textbf{R-2} & \textbf{R-L}  \\
    \midrule
    Pretrained T5  & 66.45 & 39.01 & 52.46\\
    Randomly initialized T5 & 40.07 & 20.95 & 32.42\\
    \bottomrule
    \end{tabular}
    \caption{Impact of pretraining on performance of T5-Base model on  medical dataset with {\sc Cluster2Sent} using oracle noteworthy utterance clusters}
    \label{tab:dopretrain}
\end{table}

\subsection*{Decoder Results with Oracle extracts}
We present additional quantitative results (Table~\ref{tab:rouge_scores_detailed}), 
including
(i) The ROUGE scores on the test set 
when using oracle noteworthy utterances 
with both oracle and predicted
clusters (for {\sc Cluster2Sent} models).
(ii) Two ablations on {\sc Ext2Sec}:
{\sc AllExt2Sec} uses binary classification to extract all noteworthy utterances
(not per-section), and an abstractive decoder that conditions on the section;
while {\sc Ext2SecNoCond} uses a multilabel classification based extractor 
but \emph{does not use section-conditioning} in the abstractive module. 
Both methods mostly perform worse than {\sc Ext2Sec} demonstrating the benefit of using both section-specific extraction and section-conditioning in abstractive decoder.

\begin{table*}[h!]
    \centering
    \begin{tabular}{l c c c c c c c}
    \toprule
    & \multicolumn{3}{c}{\textbf{Medical dataset}} & \hspace{0pt} & \multicolumn{3}{c}{\textbf{AMI corpus}} \\
    \midrule
    \textbf{Method} & \textbf{R-1} & \textbf{R-2} & \textbf{R-L} & & \textbf{R-1} & \textbf{R-2} & \textbf{R-L}  \\
    \midrule
        {\sc Ext2Note} (PG) & 52.95 & 27.6 & 32.87 & & 21.23 & 6.71 & 14.95\\
        {\sc Ext2Note} (T5-Small) & - & - & - & & 41.10 & 14.12 & 25.03\\
{\sc AllExt2Sec} (PG) & 50.74 & 24.33 & 32.18 & & 40.20 & 13.71 & 22.52 \\ 
{\sc AllExt2Sec} (T5-Small) & - & - & - & & 41.68 & 15.43 & 24.72 \\
{\sc Ext2SecNoCond} (PG) & 56.10 & 32.05 & 43.23 & & 42.51 & 15.71 & 23.79 \\
{\sc Ext2SecNoCond} (T5-Small) & 58.69 & 34.92 & 47.24 & & 48.14 & 18.49 & 28.23 \\
        {\sc Ext2Sec} (PG) & 61.00 & 33.64 & 45.2 & & 43.30 & 16.56 & 24.83 \\
        {\sc Ext2Sec} (T5-Small)& 62.37 & 36.39 & 49.11 & & 46.85 & 18.19 & 28.74\\
        {\sc Cluster2Sent} (PG) & 63.63 & 35.62 & 48.85 & & 51.86 & 21.86 & 31.84 \\
        {\sc Cluster2Sent} (T5-Small)&  66.50 &	38.41 &	51.73  & & 54.23 & 22.90 & 34.54 \\
        {\sc Cluster2Sent} (T5-Base)& 66.45 & 39.01 & 52.46 & & 57.42 & 24.45 & 35.70\\
        \midrule
        {\sc Cluster2Sent} (PG+clustering heuristic) & 63.12	& 35.08	& 47.96 & & 47.17 &	18.99 &	27.31 \\
        {\sc Cluster2Sent} (T5-Small+clustering heuristic)& 66.08 &	37.73 &	50.66 & & 47.53 &	19.70 &	28.95 \\
        {\sc Cluster2Sent} (T5-Base+clustering heuristic)& 65.94 &	38.26 &	51.31 & & 51.24 &	21.47 &	29.81\\
    \bottomrule
    \end{tabular}
    \caption{ROUGE scores achieved by different abstractive decoders using oracle noteworthy utterances}
    \label{tab:rouge_scores_detailed}
    \vspace{-0.3cm}
\end{table*}

\subsection*{Impact of copy mechanism}
When we do not use copy mechanism in the pointer-generator model, we observed a drop in its performance in the {\sc Cluster2Sent} setting with oracle noteworthy noteworthy utterances and clusters(Table~\ref{tab:mustusecopy}). Hence, we have used copy mechanism in all the pointer-generator models we train in this work.

\subsection*{Impact of pretraining}
When training a randomly initialized  T5-Base model on the medical dataset, even in {\sc Cluster2Sent} setting with oracle clusters, it only got a ROUGE-1 around 40 (Table~\ref{tab:dopretrain}). This is over 16 points lower than what we get by starting with off-the-shelf pretrained T5 parameters, and is even worse than {\sc Conv2Note}, highlighting the importance of pretraining.

\subsection*{Sample generated SOAP notes}

\begin{figure*}[t]
    \makebox[\linewidth]{
        \includegraphics[width=\linewidth]{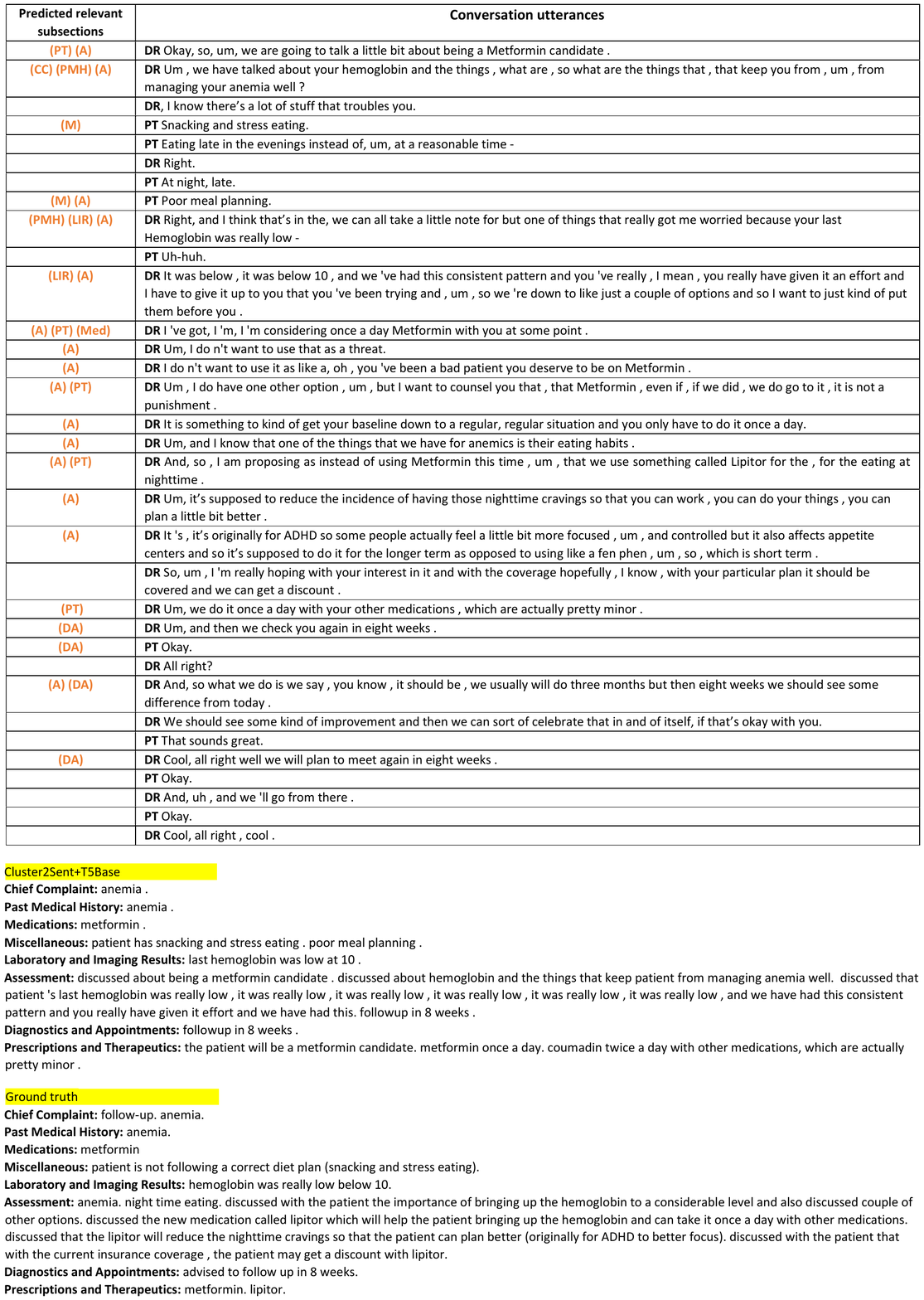}
    }
    \caption{Sample conversation (obfuscated) with SOAP note generated by the best method and the ground truth}
    \label{fig:fullconv} 
\end{figure*}

Due to privacy concerns, 
we can not publish conversations from our dataset.
Here, we present an obfuscated conversation from our test dataset, 
modified by changing sensitive content 
such as medicines, diseases, dosages (Figure~\ref{fig:fullconv}). 
We also present the SOAP note generated by our best method, as well as the ground truth.

\begin{figure}[b!]
    \includegraphics[width=0.48\textwidth]{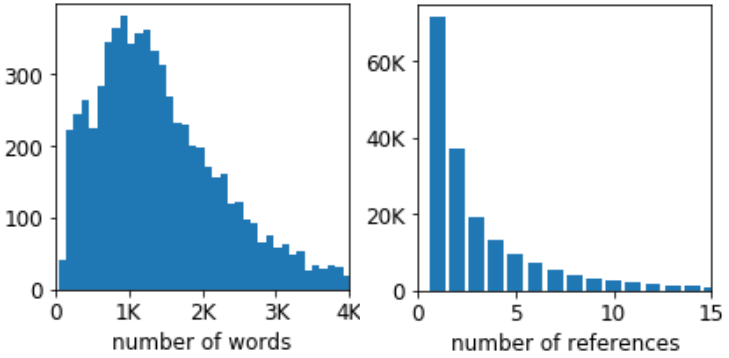}
    \caption{
Histogram of number of words in a conversation and the number of evidence utterances per summary sentence for the medical dataset}
        \label{fig:datasetstats}
\end{figure}

\subsection*{Model implementation details}
For the hierarchical LSTM classifier, we have a word embedding size of $128$ and both bidirectional LSTMs have a hidden size of $256$. For BERT-LSTM, the BERT embeddings are initialized from bert-base-uncased (768 dimensions). LSTMs in either direction have a hidden-layer of size $512$ and the entire model is optimized end-to-end with a learning-rate of 0.001. For BERT-LSTM, an input conversation is divided into chunks of 128 utterances. Due to GPU constraints, these chunks are processed one at a time.
The pointer-generator models have a word embedding size of $128$, and a hidden size of $256$ for both the encoder and the decoder. The section embeddings used in section-conditioned pointer-generator network have $32$ dimensions. During training of all pointer-generator models, the model is first trained without coverage loss~\citep{tu2016modeling} to convergence, and then trained further with coverage loss added. We tried coverage loss coefficients varying from 0.5 to 8.

The pointer-generator models were trained using Adam optimizer before coverage and using SGD after adding coverage. We tried learning rates between $10^{-4}$ and $10^{-3}$ with Adam. The next word prediction accuracy
was used as the validation criterion for early stopping while training abstractive modules, with the
exception of coverage-augmented models that used a combination of crossentropy and coverage loss. Micro-averaged AUC was used as the validation criterion for training of extractive modules.   

We employ beam search with beam size $4$ 
to decode outputs from both models. 
For the vanilla pointer-generator model 
used in {\sc Conv2Note} and {\sc Ext2Note}, 
we modified the beam search procedure 
to make sure that all the SOAP note sections 
are generated in proper order. 
We start the beam search procedure by feeding 
the header of the first section (chief complaint). 
Whenever the model predicts a section header 
as the next word and it shows up in a beam, 
we check if it is the next section to be generated.
If not, we replace it with the correct next section's header. 
Any \texttt{end-of-summary} tokens generated 
before all the sections have been produced are also replaced similarly. 
Note that producing all sections simply means 
that the headers for each section have to be generated,
and a section can be left empty by starting 
the next section immediately after generating the previous header. 
The decoding length for beam search is constrained to be between $5^{th}$ and $95^{th}$ percentile of the target sequence length distribution, calculated on the training set.

\subsection*{Simulating ASR Errors}
We simulate ASR errors at any given percentage rate by randomly selecting the percentage of the words in the conversation and replacing them with phonetically similar words. To reduce the search space of possible candidates for each word, we use the \url{suggest()} function taken from the Pyenchant\footnote{https://pypi.org/project/pyenchant/} library that provides auto-correct suggestions for the input word. 
Each suggestion is then passed through the Refined SoundEx algorithm to find the phonetic distance between the original and the suggested word. We use the pyphonetics\footnote{https://pypi.org/project/pyphonetics/} package for a python implementation of this algorithm. For our final candidate list, we choose words that are at phonetic distance of $1$ from the original word. Finally, a candidate is chosen at random from this list to replace the original.

\begin{table*}[t]
    \centering
\resizebox{0.8\linewidth}{!}{
\begin{tabular}{lccccccc}
\toprule
    & \multicolumn{3}{c}{\textbf{Medical conversations}} & \hspace{10pt} & \multicolumn{3}{c}{\textbf{AMI corpus}} \\
\midrule
Count & C2N & C2S-P & C2S-T &    								& C2N & C2S-P & C2S-T \\
\midrule
Total sentences		   	  & 956 & 1268 & 1277 &    				& 414 &  358  &  381  \\
Repetitive			  	  & 96 & 127 & 147 &    				& 213 &  14  &  14  \\
Incoherent 			  	  & 162 & 158 & 58 &    				& 9 &  134  &  27  \\
True statements 				  & 587 & 848 & 931 &    		& 89 &  103  &  227  \\
False statements 				  & 100 & 116 & 125 &    		& 71 &  75  &  68  \\
Truthfulness undecided 				  & 11 & 19 & 16 &    		& 32 &  32  &  45  \\
Irrelevant 				  & 25 & 34 & 24 &    					& 14 &  15  &  2  \\
Under incorrect section 			  & 56 & 42 & 39 &    		& 4 &  2  &  18  \\
\bottomrule
\end{tabular}
}
    \caption{Number of sentences produced by different methods that were judged to have different listed characteristics by human raters. C2N:{\sc Conv2Note}, C2S-P:{\sc Cluster2Sent} with pointer-generator, C2S-T:{\sc Cluster2Sent} with T5-base. BERT-LSTM used for medical dataset, hierarchical-LSTM used for AMI corpus.}
    \label{tab:humaneval_detailed_results}
\end{table*}

\begin{table*}[h]
    \centering
\resizebox{0.85\linewidth}{!}{
\begin{tabular}{lccccc}
\toprule
\textbf{Subsection} &  \textbf{ROUGE-1} &  \textbf{ROUGE-2} &  \textbf{ROUGE-L} &  \textbf{N} & \textbf{L} \\
\midrule
chief complaint                &   44.34 &   28.12 &   43.59 &         592 &                    11.46 \\
review of systems              &   46.88 &   28.35 &   43.28 &         514 &                    29.24 \\
past medical history           &   53.48 &   37.70 &   51.80 &         547 &                    17.81 \\
past surgical history          &   58.44 &   43.08 &   57.04 &         230 &                    10.36 \\
family medical history         &   51.94 &   36.49 &   50.13 &          72 &                    16.14 \\
social history                 &   57.72 &   37.82 &   56.30 &          97 &                    10.33 \\
medications                    &   49.56 &   23.53 &   47.64 &         549 &                    15.28 \\
allergies                      &   39.29 &    6.63 &   38.32 &          21 &                     8.57 \\
miscellaneous                  &   28.87 &   11.61 &   24.90 &         415 &                    34.44 \\
immunizations                  &   55.95 &   27.49 &   54.81 &          25 &                     7.32 \\
laboratory and imaging results &   58.36 &   41.18 &   55.11 &         448 &                    19.37 \\
assessment                     &   39.01 &   15.31 &   25.35 &         570 &                   132.41 \\
diagnostics and appointments   &   52.85 &   35.70 &   50.43 &         488 &                    17.67 \\
prescriptions and therapeutics &   50.53 &   33.51 &   48.10 &         446 &                    18.73 \\
healthcare complaints          &   30.11 &   15.79 &   29.57 &          43 &                    16.74 \\
\bottomrule
\end{tabular}
}
    \caption{Average ROUGE scores (from {\sc Cluster2Sent}  T5Base+BLSTM) for each section of SOAP note (N-number of test datapoints with the section populated, L-average number of words in ground truth)}
    \label{tab:sectionwise_results}
\end{table*}

\begin{table*}[h!]
\small
    \centering
\resizebox{0.95\linewidth}{!}{          
\begin{tabular}{lrrrrrrr}
\toprule
\textbf{Section} &  \textbf{Base rate(\%)} &  \textbf{Precision} &  \textbf{Recall} &     \textbf{F1} &  \textbf{Accuracy} &    \textbf{AUC}  \\
\midrule
chief complaint                &  3.12 &      34.71 &   33.93 &  34.31 &     95.95 &  86.81\\
review of systems              &  5.10 &      51.35 &   51.82 &  51.58 &     95.04 &  93.12\\
past medical history           &  3.41 &      36.00 &   36.52 &  36.26 &     95.63 &  88.00\\
past surgical history          &  0.99 &      33.80 &   34.50 &  34.14 &     98.68 &  93.74\\
family medical history         &  0.31 &      52.31 &   45.25 &  48.53 &     99.70 &  99.23\\
social history                 &  0.53 &      59.81 &   54.87 &  57.23 &     99.56 &  95.41\\
medications                    &  4.45 &      51.66 &   49.13 &  50.36 &     95.69 &  92.02\\
allergies                      &  0.16 &      30.86 &   12.44 &  17.73 &     99.82 &  89.46\\
miscellaneous                  &  3.71 &      24.06 &   16.17 &  19.34 &     95.00 &  80.05\\
immunizations                  &  0.05 &      63.64 &   64.62 &  64.12 &     99.96 &  97.63\\
laboratory and imaging results &  2.46 &      50.00 &   55.15 &  52.45 &     97.54 &  93.84\\
assessment                     & 14.19 &      38.09 &   42.01 &  39.96 &     82.08 &  76.89\\
diagnostics and appointments   &  2.10 &      55.60 &   40.16 &  46.63 &     98.07 &  94.22\\
prescriptions and therapeutics &  3.11 &      41.28 &   38.43 &  39.81 &     96.39 &  92.40\\
healthcare complaints          &  0.25 &      20.47 &   21.90 &  21.17 &     99.60 &  85.93\\
\bottomrule
\end{tabular}
}
    \caption{Performance of BERT-LSTM on extracting noteworthy utterances for various SOAP sections}
    \label{tab:extractor_performance_sectionwise}
\end{table*}

\subsection*{More Experimental Details}
We trained models on multiple Nvidia \texttt{Quadro RTX 8000}, \texttt{RTX 2080Ti} and \texttt{V100} GPUs.
The extractive modules were evaluated
using standard classification metrics
from \texttt{scikit-learn}~\footnote{\url{https://scikit-learn.org}}
and quality of summaries were evaluated
using ROUGE scores calculated with the \texttt{pyrouge} Python package~\footnote{\url{https://pypi.org/project/pyrouge}} which is a wrapper around the \texttt{ROUGE-1.5.5} Perl script.

\end{document}